\documentclass{article}
\usepackage{graphicx}
\usepackage{amsmath}

\newtheorem{example}{Example}

\def\KB{\textit{KB}}
\def\AF{\texttt{AF}}
\def\Args{\texttt{Args}}
\def\Att{\texttt{Att}}
\def\Def{\texttt{Def}}
\newcommand{\tuple}[1]{\langle#1\rangle}
\newcommand{\cond}[1]{\emph{(#1)}}

\title{Abduction and Argumentation for Explainable Machine Learning: A Position Survey\thanks{A version of this paper will appear in ``Dinh Phung, Claude Sammut, and Geoffrey I. Webb (eds). \emph{Encyclopedia of Machine Learning and Data Science, 3rd edition.}, in preparation.''}}

\author{
Antonis Kakas\\
Department of Computer Science,\\ 
University of Cyprus, Cyprus\\
antonis@ucy.ac.cy
\and
Loizos Michael\\ 
Open University of Cyprus \&\\
Research Center on Interactive Media,\\
Smart Systems, and Emerging Technologies\\
loizos@ouc.ac.cy
}

\begin{document}

\maketitle

\begin{abstract}
This paper presents Abduction and Argumentation as two principled forms for reasoning, and fleshes out the fundamental role that they can play within Machine Learning. It reviews the state-of-the-art work over the past few decades on the link of these two reasoning forms with machine learning work, and from this it elaborates on how the explanation-generating role of Abduction and Argumentation makes them naturally-fitting mechanisms for the development of Explainable Machine Learning and AI systems. Abduction contributes towards this goal by facilitating learning through the transformation, preparation, and homogenization of data. Argumentation, as a conservative extension of classical deductive reasoning, offers a flexible prediction and coverage mechanism for learning --- an associated target language for learned knowledge --- that explicitly acknowledges the need to deal, in the context of learning, with uncertain, incomplete and inconsistent data that are incompatible with any classically-represented logical theory.
\end{abstract}


\section{Introduction}%
\label{Section: Introduction}

Abduction and Argumentation are two forms of inference where conclusions are drawn according to an underlying theory. Typically, abduction aims to draw an explanation for a set of observations, while argumentation aims to give reasons, or arguments, that support a conclusion against other conflicting conclusions.

Abduction is sometimes described as ``deduction in reverse'', whereby given a rule ``A follows from B'' and the observed result ``A'', we infer that the condition ``B'' of the rule (may) hold. More generally, in the context of a logic-based setting, given a set of sentences representing a theory $T$ that models a domain of interest, and a sentence representing an observation $O$, abduction returns a set of sentences representing an abductive explanation $H$ for $O$, such that:

\begin{enumerate}
\item[{1.}] $T \cup H\;\models \;O,$
\item[{2.}] $T \cup H$ is consistent,
\end{enumerate}

\noindent where $\models$ denotes the deductive (or other) logical entailment relation of the formal logic used in the representation of our theory, and consistency also refers to the corresponding notion in this formal logic. Several abductive explanations can exist for the same observation, and in many cases additional requirements may be imposed, such as for example minimality, for an explanation to be admissible.

Argumentation is concerned with supporting a claim (e.g., a belief or a decision) based on some premises and an argument that links these premises to the claim. An arguments in support of a claim is expected to be acceptable or valid, in the sense of being able to defend itself against all other arguments that are in conflict with it, i.e., counter-arguments that are challenging the supporting argument. 

In a formal setting, argumentation takes place within a given argumentation framework\footnote{There are different variations of the definition of an argumentation framework in the literature \cite{Dung1995,Modgil2009}.} $\AF = \tuple{\Args, \Att, \Def}$, where $\Args$ is a set of arguments, $\Att$ is a binary attack relation on $\Args$, and $\Def$ is a binary defense (or defeat) relation on $\Args$. The attack relation specifies when one argument is a counter-argument to (i.e., opposing or challenging) another argument, while the defense relation captures the notion that an argument is sufficiently strong to defend against another (opposing or challenging) argument. Given an argumentation framework $\AF$, a subset of arguments, $\Delta$, is acceptable in $\AF$, iff:

\begin{enumerate}
\item[\cond{1}] $\Delta$ is conflict-free; i.e., it does not include arguments that attack each other.
\item[\cond{2}] $\Delta$ defends against every other subset of arguments $A$ that attacks it\footnote{More generally, a subset $\Delta$ of arguments is acceptable when it can render all its counter-arguments (or counter subsets of arguments) non-acceptable.}.
\end{enumerate}

As this definition indicates, the process of argumentation does not in general produce a single supporting argument $a_0$ for a desired claim, but its aim is to form a coalition, $\Delta$, of $a_0$ with other arguments so that $\Delta$ can defend $a_0$ against other arguments that would undermine it in some way; e.g., by questioning its premises, or indeed the appropriateness of the argument's link between its premises and the claim (the argument scheme) used to construct $a_0$.

Given the above, a claim $\phi$ is: credulously (or possibly) entailed by $\AF$ iff there exists an acceptable coalition $\Delta$ under $\AF$ that supports $\phi$; skeptically (or strongly) entailed by $\AF$ iff it is credulously entailed by $\AF$ and no other claim $\psi$ that is incompatible (or in conflict) with $\phi$ (e.g., $\psi = \neg \phi$) is credulously entailed by $\AF$. In effect, deductive entailment is replaced by the more general\footnote{Deduction in its classical formulation can be shown to be a limiting case of this form of argumentation-based logic \cite{AL_StudiaLogic}.} case of logical reasoning via argumentation. We will see below the significance of this new perspective on reasoning, in relation to the notion of coverage and prediction in logic-based or symbolic learning.

\subsection{The Link Between Abduction and Argumentation}

Abduction and Argumentation can both be seen as processes for generating explanations either for a given observation as in the case of abduction or for a conclusion (claim or decision) in the case of argumentation. Explanations under abduction are in terms of underlying (theoretical or non-observable) hypotheses, whereas explanations under argumentation are in terms of arguments (among a set of known ones) that provide justified reasons for a conclusion to hold.

Abduction and Argumentation are closely linked. Argumentation can be viewed under the abduction lens as a process of explaining a claim by the set of premises that the supporting argument is based on, linking these to its claim. Furthermore, argumentation can offer a more complete explanation for believing or accepting a claim by providing also reasons against alternative claims, as these form counter-arguments to the chosen claim, which are defended against by the acceptable argument supporting the claim. In other words, these defense arguments form an extension of the explanation for the claim by also giving the reasons why to accept or prefer this claim over other competing alternatives. In the context of abduction, this allows argumentation to be used to provide (and explain) the reasons to prefer one abductive explanation over another abductive explanation for the same observation, as such alternative explanations are typically considered as competing alternatives. Thus, argumentation can help abduction realize its often quoted requirement of ``inference to the best explanation''.

Another way to understand this link between abduction and argumentation is to notice that abduction can be performed with respect to a theory $T$ that is, in fact, an argumentation framework, i.e., $T=\AF$, and that the entailment relation $\models$ is that of credulous or sceptical entailment under argumentation. With this link abduction can help enhance argumentation in cases where when forming arguments, and defending these against their counter-arguments, it is necessary to make assumptions about missing supporting premises or about other properties that are not known explicitly, but are needed to form a possible defense against counter-arguments. Argumentation then depends on abduction to produce a set of underlying hypotheses that would help the formation of a properly justified argument \cite{Arg_Abd_AAMAS03,Abd_In_Arg_Sakama}. These abductive hypotheses become, thus, part of the argument --- the reason --- for supporting its claim.
 
On the other hand, we can obtain argumentation via abduction by associating to any argument an abducible assumption that carries the burden of the use of the argument such that an argument is attacked by attacking its corresponding assumption, and hence an acceptable set of arguments $\Delta$ will correspond to a set of the abducible assumptions linked to the arguments in $\Delta$ \cite{ABA}.


\section{Motivation \& Background: Link to Learning}%
\label{Section: Motivation and Background}

Reasoning and Learning have a synergistic inter-dependence: learning produces the knowledge that is assumed as given when reasoning; reasoning draws inferences that provide the inductive bias that is assumed as given when learning. How can we exploit this synergy, particularly with the reasoning processes of abduction and argumentation, so that we can enhance the learning process?

The acknowledgement of the existence of this inter-dependence becomes particularly important in the backdrop of a major recent development in Machine Learning and AI: the emergence of the need for explainability in Machine Learning (and Deep Learning, in particular), where decisions taken on the basis of learned models need to be transparent and comprehensible to human users \cite{XAI_BARREDOARRIETA2020,XAI_guidotti2018survey}, and where neural-symbolic integration can help develop such explainable systems, based both on lower-level sensory data and higher-level cognitive data. Although the detailed operation of a learned hypothesis may be unknown, it should be possible to explain its inferences at a level that is cognitively-compatible with the intended users or consumers of those inferences. Explanations should not require the users to have technological knowledge and should be offered at the high-level conceptual language of the application as used by the application experts and/or users. Abduction and argumentation, as mechanisms that generate explanations, can support directly the realization of this need for Explainable ML, achieving explainability by design of the way they reason over a learned theory. 

\subsection{Representation and Reasoning for Learned Knowledge}

How are we to reason with a learned theory, e.g., to predict the properties of new cases not included in the learning training data? Argumentation (as a conservative extension of deduction in the face of conflicting information) can replace deductive reasoning as a more informative and flexible notion of coverage in learning. The flexibility afforded by argumentation makes it a natural target language for learning, as it acknowledges the default nature of inductively learned knowledge. To demonstrate this natural connection, we consider the celebrated example of Pierce \cite{Pierce} on the color of beans in a bag. Having observed that all beans drawn out of this bag so far are white, and having formed the inductive generalization that ``All beans in this bag are white'', we face a problem when we draw a non-white bean from the bag. Is the generalization not useful and should be abandoned altogether? If we keep the generalization, how would we reason with it, especially since formal classical deductive logic would not be appropriate given the observed counter-example of the non-white bean?

As early as in the 18th century, Hume \cite{Hume_Treatise} pointed out that inductive generalization that is universal and absolute runs into logical difficulties as we cannot be sure that a future case will not contradict the generalization. Understanding the logic of generalization has come to be known in philosophy as the ``the problem of induction'' \cite{SEP_Hume_Induction}. By the very nature of the task, learning leads to information that in general cannot be absolute, and may contain some element of uncertainty or incompleteness about the underlying knowledge that we are trying to learn. Only in special and limiting cases we arrive at an absolute and complete understanding of what we are trying to learn. 

One way to address this foundational question on induction is to consider a target learning language based on argumentative and/or abductive reasoning. Then, inductively-produced rules associating different concepts can be ``abductive rules'' with a missing element which we need to assume when reasoning with the learned theory. Hence, the learned generalization from the beans in the bag can be represented by ``All \emph{normal} beans from this bag are white'', where the condition of normality\footnote{Note here assumption of normality refers to a theoretical or non-observable concept, as is the usual case with abductive reasoning \cite{FK00}.} can be abductively assumed for any new bean drawn from the bag. Thus, a new bean will be predicted to be white based on the assumption that it is a normal bean. If indeed it is observed to be white, this would be abductively explained by the assumption that it is a normal bean. If on the other hand, we observe a black bean drawn from the bag, our theory can explain its black color by the assumption that it is not a normal bean.

In time we may be able to learn some properties of these unknown abducible assumptions in our generalizations (learned rules), which would allow us to be more informed about what these missing abductive assumptions are, or deter us from making certain such assumptions; e.g., if the bean feels small then this is not a normal bean in this bag. In particular, it may be possible to start learning other generalizations from the available data that would be usefully integrated together with our earlier generalizations to give an enhanced prediction capability of the integrated learned theory. For example, we may form the inductive generalization that ``all small beans from this bag are black'', complementing the original generalization that ``all beans from this bag are white''.

In this way we are naturally led to view the inductive generalizations as providing arguments for the various possible conclusions, rather than strict or absolute rule associations between the concepts involved. For example, the generalization of ``all beans from this bag are white'' can be interpreted as providing an argument $a_1$ supporting the claim that a bean is white based on the premise that it was taken from this bag. Similarly, an inductive generalization of ``all small beans from this bag are black'' provides an argument $a_2$ supporting the claim that a bean is black based now on the premise that it is a small bean from this bag.

Argumentative reasoning will then juxtapose such arguments in a debate to try to reach a conclusion that is supported by an acceptable argument. In this debate, arguments which are (considered) relatively stronger will win, and their claims will be skeptically predicted. In our example above, if a new bean taken from the bag is small then argument $a_2$ would win the debate provided that we consider $a_2$ to be stronger than $a_1$; this relative strength could be naturally justified because the premises of $a_2$ are more specific than those of $a_1$.

If no such stronger arguments exist we would typically not be able to reach a clear conclusion, and we would have a dilemma where different and conflicting conclusions could each be supported acceptably by their own arguments. For example, we may have also learned the argument $a_3$ that wrinkled beans from this bag are green. What would we then predict in a new case where the bean is small and wrinkled? As our learned theory stands, we would have three relevant arguments one claiming white, one black, and the third green as the color of this new bean. The first argument will be defeated by (each of) the other two stronger arguments (due to their specificity over $a_1$). But between $a_2$ and $a_3$ the process of argumentation will not produce a clear winner, and argumentation will offer two credulous predictions, each one accompanied by an explanation.

Although this may not seem to be a completely satisfactory result, the view of learning as a process of producing arguments opens up a new view of reasoning with learned knowledge, where although a final firm decision might not be reached, reasoning is at least able to provide reasons or explanations for concluding one way or another --- by presenting (in a suitable way) the arguments that support each of the different possibilities. This could help point towards where we should concentrate any further learning, e.g., the type of further training data that we should collect. Also, potentially, the resolution of such reasoning dilemmas could be offloaded to human experts, embracing a more symbiotic relation between humans and machines working together to reach a decision.

The aforementioned ideas are made precise in our upcoming proposal for a framework on \emph{cognitively-explainable learning} \cite{CEL}, where the use of argumentation as a target learning language is shown to naturally facilitate the formalization of dilemmas in the broader context of building explainable AI systems with learning guarantees. This framework naturally accommodates the consideration of whether the learned argumentation theory is cognitively compatible with the particular human who is consuming the arguments, and who is collaborating with the machine to resolve the dilemmas and draw appropriate inferences.

The feasibility of this symbiosis is supported by decades of work in Cognitive Science and Philosophy, which has confirmed the primacy of argumentation in human reasoning, through a multitude of empirical studies within the area of Cognitive Psychology (e.g., \cite{mercier}). These empirical studies differ both in the type of experiments carried out, but also in the way that argumentation is formulated for the study. Thus, the underlying position of the primacy of argumentation in human reasoning is solidly confirmed via these different perspectives.

There has also been an extensive study of formulating the many different formal frameworks for non-monotonic --- sometimes called commonsense --- reasoning in AI, in terms of argumentation, showing the universality of argumentation as a logical framework. With the emphasis (and return) on Human-Centric and Explainable AI, the use of argumentation as a target language for machine learning could help achieve a seamless integration of AI systems in human lives, by supporting cognitively more comprehensible and understandable learned theories. 

\subsection{Guiding Learning through Reasons and Explanations}

At a technical level, the explanation-generating nature of abduction and argumentation help utilize the current knowledge to facilitate the further and more complete learning. Abduction and argumentation fulfil this function in distinct ways. With abduction explanations provide a form of translation, rationalization, or homogenization of the training data into a common underlying level at which learning is carried out. With argumentation explanations about the reasons why training data are classified in a certain way orients and focuses the learning process.

For abduction, one typically considers that the current knowledge is relatively progressed with a good model, where the incompleteness of the knowledge is isolated to an underlying level that in general is not directly observable or available from the environment. Abduction then generates, in its explanations, new information at this underlying level not hitherto contained in the current theory which can then facilitate the learning of new knowledge in terms of associations between the abducibles. As a case in point, in the context of neural-symbolic integration --- an approach that can help develop coherent systems that cope with lower-level sensory data and higher-level cognitive data and enhance, thus, existing deep learning theories with an explainable front-end component --- and assuming that the symbolic part of the architecture is mature enough in terms of the knowledge it utilizes, the abductive explanation process of rationalization can help by translating high-level training data into training data at the lower level of a neural module. Therefore, while the neural module might not directly have access to labels that explicitly correspond to its inputs, the process of abduction can translate the high-level label of those inputs into lower-level supervision signals for the neural module.

For argumentation, the current knowledge may not be as developed (it could even be empty), and could still be full of conflicting possibilities for many cases of the training data. Indeed, a problem domain may be inherently one where we cannot (yet) get the same level of isolation of incompleteness of a theory to an underlying level, and we need to tolerate or accommodate the uncertainty of multiple interpretations of cases. By recognizing such dilemma cases (where more than one mutually incompatible prediction would be credulously entailed), argumentation helps to indicate how the current knowledge is to be extended or revised. Knowledge that presents a dilemma on interpretations suggests to the learning process to focus on those training data that would help resolve the dilemma. For example, faced with the dilemma of what color a small wrinkled bean is, learning might seek to identify the conditions under which one argument among those supporting the two choices (of black or green) is stronger.

This means that within an argumentation-based learning framework we want to learn not only object-level arguments, but also meta-knowledge in the form of a relative strength or preference between arguments. This meta-knowledge feeds into the defense relation between arguments, which in turn influences the credulous or sceptical predictions that we can draw from the learned theory. Note that such learned meta-level preferences can be conditional on properties of our learning samples, drawing in effect non-linear separation or classification lines in the learning space.


\section{Structure of the (Machine) Learning Task}%
\label{Section: Structure of the Learning Task}

\begin{figure}[t]
\centering
\includegraphics[width=\textwidth]{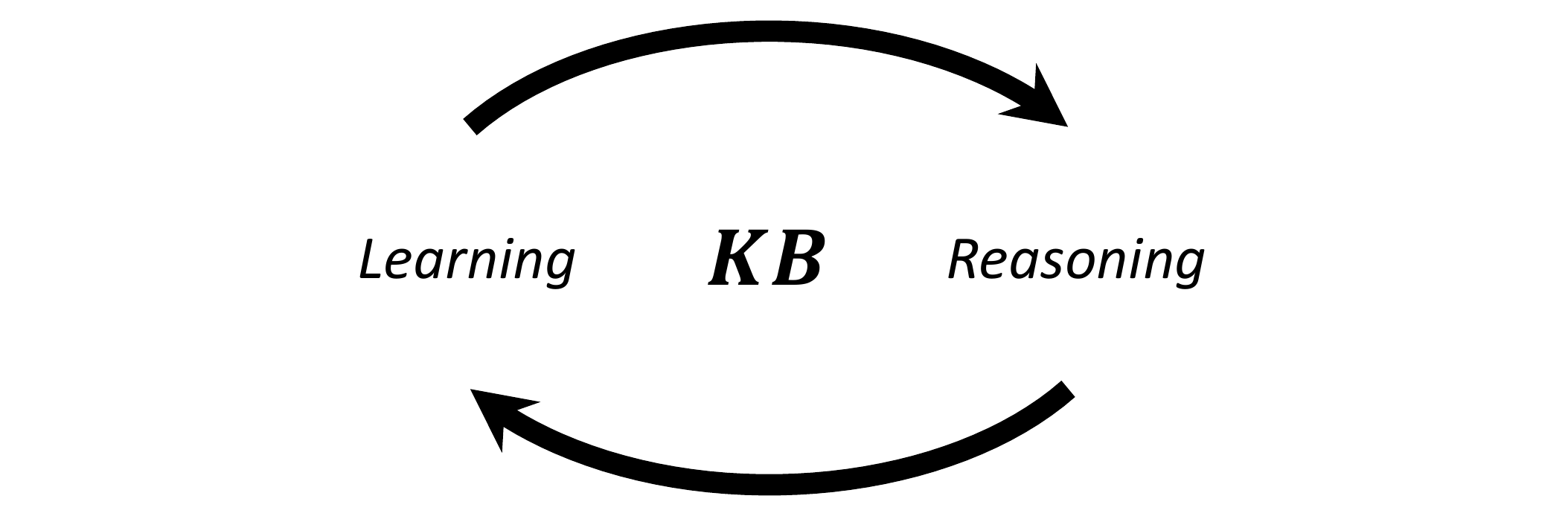}
\caption{The cycle of reasoning and learning in a process of knowledge development. Learning generates new knowledge $\KB$ that feeds into reasoning about new problems. Reasoning about the learning data under the current knowledge $\KB$ feeds into the learning process to generate additional knowledge.}
\label{Figure: Reasoning-Learning Cycle}
\end{figure}

The synergistic inter-dependence between Reasoning and Learning naturally leads to a continuous cycle of interaction between the two processes (see Figure~\ref{Figure: Reasoning-Learning Cycle}), providing for the incremental development of the knowledge $\KB$ about the domain of interest. Such an incremental development of the knowledge is necessary when the various knowledge parts (e.g., rules) build on each other, as it ensures that the learning guarantees apply equally for all learned parts \cite{SLAP}.

Due to the different nature of abductive and argumentative reasoning, their function in the reasoning-learning cycle is different in its specifics. Both processes, however, have an important role to play in both parts of the cycle: outside learning, they act as explanation-generation mechanisms; and during learning, they utilize their generated explanations to facilitate learning.

\medskip

\noindent \textbf{Abduction} needs prior knowledge $\KB$ to operate, and explanations are according to $\KB$. Abductive reasoning influences learning by preparing the learning data and transferring the learning to the underlying level of explanations.

\begin{description}

\item[Outside Learning:] Abduction uses $\KB$ to reason on new cases of interest by providing explanations of the observed status of the new cases.

\item[During Learning:] Abductive reasoning \emph{rationalizes}, according to $\KB$, the training data, thus homogenizing or normalizing the data at an underlying level in $\KB$ in a way that can help the inductive learning process. Abductive reasoning also \emph{imputes} missing information in the background attributes/features of the training data, so that under such abductive hypotheses the coverage of a candidate learned theory is improved.

\end{description}

\noindent \textbf{Argumentation} reasons with, or analyzes, conflicts in the current knowledge $\KB$. As such, it offers a flexible notion of covering for learning that can help evaluate, understand, or isolate, during the learning process, the inadequacy of the current candidates for the knowledge to be learned, and incrementally revise the learned theory $\KB$ through a progressive reduction or resolution of conflicts.

\begin{description}

\item[Outside Learning:] Argumentation reasons on new cases of interest to predict their status under the learned theory $\KB$, to identify dilemmas among alternative predictions, and to provide reasons supporting each alternative.

\item[During Learning:] Argumentative reasoning as a covering notion \emph{interprets} the learning data with reasons for the different possible alternatives for their clarification/interpretation, and makes explicit the dilemmas that exist, according to $\KB$, among those alternatives. These dilemmas help to concentrate on the part of the information that is more relevant and focus the further learning effort on subcases where conflicts continue to exist.

\end{description}

We continue below to expand on the link of the reasoning modes of argumentation and abduction to the learning process, and ground this link on specific approaches as found in the main relevant literature.


\subsection{Abduction in Machine Learning}

Logic-based abduction generates a set of hypotheses, \textit{H}, that when added to the given theory \textit{T} this would logically entail a given set of observations \textit{O}. This basic formalization as it stands, does not fully capture the explanatory nature of the abductive hypothesis \textit{H}, in the sense, that it necessarily conveys some reason for why the observations hold. Typically, we want to express the abducive explanations in a way that they convey some ``deeper'' reason for which the observations must hold according to the theory \textit{T}, e.g., explanations could be in terms of hypotheses that would causally imply the observations through the given theory. Therefore we would normally specify a ``level'' in the theory at which the abductive explanations are expressed. One way to do this is simply to specify the vocabulary and language for expressing explanations restricting this to a special preassigned, domain-specific class of sentences called \textit{abducible}.

\subsubsection{Abducing Background Knowledge on the Training Data}

In the context of Machine Learning the simplest case of this abducible explanation level is the level of the background knowledge associated with the training data. For example, the attribute or feature description of the training examples may be incomplete for some of these examples. Abductive reasoning can be used to fill in these gaps with informed hypotheses for their value thus helping to prepare the training data for the learning process.

Abductive concept learning (ACL) \cite{ACL} is a learning framework that allows us to learn from such incomplete information and to later be able to classify new cases that again could be incompletely specified. Under ACL, we learn abductive theories, $\langle$\textit{T}, IC$\rangle$, that contain, together with a set of (logic program) rules, in \textit{T}, for the concept(s) to be learned, integrity constraints, in \textit{IC}, in the form of general clauses. These are used to constrain the abductive hypotheses that we can form on the background missing data in classifying new cases according to the learned theory.

The semantics of ACL require that for every positive training example, there must exist in the learned theory an abductive explanation and the collection of all such explanations for all the positive examples must be consistent with each other. For negative training examples, it is required that, within the learned theory, no abductive explanation exists for any of them. We illustrate ACL with a simple example.

\begin{example}
Suppose we want to learn the concept \textit{father} from the following given training examples:

\begin{itemize}
\item[] \textit{E}$^{+}$ = \{\textit{father}(\textit{john}, \textit{mary}), \textit{father}(\textit{david}, \textit{steve})\},
\item[] \textit{E}$^{-}$ = \{\textit{father}(\textit{kathy}, \textit{ellen}), \textit{father}(\textit{john}, \textit{steve})\}.
\end{itemize}

\noindent with background knowledge $\KB$ given by:

\begin{itemize}
\item[] \textit{\KB} = \{\textit{parent}(\textit{john}, \textit{mary}), \textit{male}(\textit{john}), \textit{parent}(\textit{david}, \textit{steve}), \textit{parent}(\textit{kathy}, \textit{ellen}), \textit{female}(\textit{kathy})\}.
\end{itemize}

\noindent Notice that the background knowledge that we have on \textit{male} and \textit{female} is incomplete and assumptions on these form the abducibles in our problem.

\noindent A possible abductive theory, $\langle$\textit{T}, \textit{IC}$\rangle$ learned by ACL would consist of

\begin{itemize}
\item[] \textit{T} = \{\textit{father}(\textit{X}, \textit{Y}) \quad  $\leftarrow$ \quad \textit{parent}(\textit{X}, \textit{Y}), \textit{male}(\textit{X})\},
\item[] \textit{IC} = \{\textit{false} \quad $\leftarrow$ \quad \textit{male}(\textit{X}), \textit{female}(\textit{X})\}.
\end{itemize}

Despite the fact that the background theory is incomplete (in its abducible predicates), ACL can find an appropriate solution to the learning problem by suitably extending the background knowledge and allowing abductive hypotheses, such as that of \textit{male}(\textit{david}), in order to cover the positive example of \textit{father}(\textit{david}, \textit{steve}). Note that the learned theory without the integrity constraint that restricts the freedom in drawing abductive hypotheses would not be a solution, because there would exist, in the learned theory, an abductive explanation for the negative example \textit{father}(\textit{kathy}, \textit{ellen}), namely \textit{male}(\textit{kathy}), i.e., this negative example would be covered as positive. This explanation is prohibited in the complete theory by the learned constraint together with the fact \textit{female}(\textit{kathy}). 
\end{example}

The learning algorithm and system for ACL is based on a decomposition of this problem into two subproblems: (1) learning the rules in \textit{T} together with appropriate explanations in \textit{IC} for the training examples and (2) learning integrity constraints driven by the explanations generated in the first part. This decomposition allows ACL to be developed by combining the two ILP settings of explanatory (predictive) learning and confirmatory (descriptive) learning. In fact, the first subproblem can be seen as a problem of learning from entailment, while the second subproblem as a problem of learning from interpretations.

An important application of ACL is that of Multiple Predicate Learning (MPL) \cite{MPL}, where each predicate is required to be learned from the incomplete data for the other predicates. Here the abductive reasoning can be used to suitably connect and integrate the learning of the different predicates by generating hypotheses on one predicate on which another predicate depends. This can help to overcome some of the nonlocality difficulties of MPL, such as order-dependence and global consistency of the learned theory.

\subsubsection{Abduction as a Case of Learning}

In some cases, when our knowledge already contains a detailed description of the problem domain and the incompleteness of its model is isolated in its abducibles, the generation of an abductive explanation/hypothesis from the given set of observations can be considered as a form of learning. This is because the abductive explanation forms a genuine new piece of information that is not already contained in (or derived from) our current knowledge and hence we have learned the underlying reasons or why, things are as observed. In other words, abductive reasoning is equated to a form of learning, sometimes referred to as abductive learning: we have learned the reason(s), according to our current model, for given observational data. In some applications of this form of abductive learning, see, e.g., \cite{AIDS_abduction,sato2002statistical,TAMADDONINEZHAD2013225}, there can be many such possible abductive explanations specific to the observational data and we then use a probabilistic analysis to extract information out of this multitude of explanations.

In general, though, this form of abductive learning is weak in the sense that its generalization effect, if at all present, is very limited. This is because abduction always needs to refer to some current given theory, from which the explanatory hypotheses are drawn, and thus the generalizing power of abduction is restricted as the basic underlying model of our domain remains unchanged. In other words, the ``learned'' abductive explanations are already contained in our theory and the possibility to apply these to other cases, other than on the observations that have generated the explanations, is very limited.

\subsubsection{Cycle of Abduction and Induction in Learning}

\begin{figure}[t]
\centering
\includegraphics[width=\textwidth]{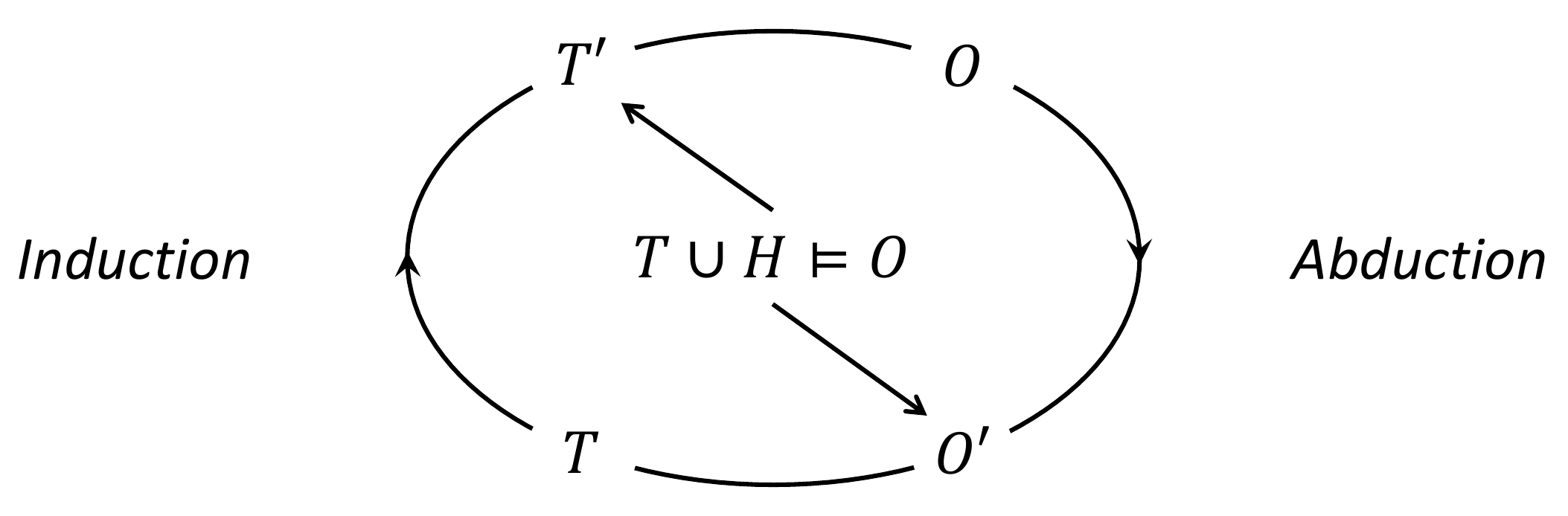}
\caption{The cycle of abductive and inductive knowledge development \cite{FK00}. The cycle is governed by the ``equation'' $T \cup H \models O$, where \textit{T} is the current knowledge, \textit{O}~the observations triggering theory development, and \textit{H} the new knowledge generated. On the left-hand side we have induction, its output feeding into the theory \textit{T} for later use by abduction on the right; the abductive output in turn feeds observational or training data \textit{O}$^{\prime}$ for use by induction, and so on.} 
\label{Figure: Abduction-Induction Cycle}
\end{figure}

To overcome this limitation, we need a synthesis with induction or some other, possibly  non-logical, process of learning, where the individual explanations for several different cases of observation are generalized and hence can be applied to genuinely-new cases. Several approaches for synthesizing abduction and induction within the process of learning have been developed, e.g., \cite{AILP_Ade_Denecker,dimopoulos1996abduction,FK00,MuggBryant,yamamoto}. These approaches aim to develop techniques for knowledge intensive learning with complex background theories. Conceptually, all these proposals fall under the same simple model of actively integrating abduction and induction in a cycle where abduction through its explanations helps to prepare the training data for induction. In general, such a \textit{cycle of integration} of abduction and induction \cite{FK00} allows the incremental development of the theory \textit{T} describing our knowledge on a domain (Figure~\ref{Figure: Abduction-Induction Cycle}).

By treating the given training data for learning as observations, abduction is used to transform (and in some sense normalize) the data to information on the abducible predicates. Then, induction takes this as input and tries to generalize this information to new knowledge on the abducible predicates now treating these as observable predicates for its own purposes. The cycle can then be repeated by adding the learned information on the abducibles back in the model as new partial information on the incomplete abducible predicates. This will affect the abductive explanations of new observations to be used again in a subsequent phase of induction. 

A simple example, adapted from \cite{Ray1}, that illustrates this cycle of integration of abduction and induction is as follows (see also \cite{dimopoulos1996abduction} for further examples):

\begin{example}
Suppose that our current model, \textit{T}, contains the following rule and background facts:

\begin{itemize}
\item[] \textit{sad(X) \quad  $\leftarrow$ \quad  tired(X), poor(X)}

\textit{tired(oli), tired(ale), tired(kr)} 

\textit{academic(oli), academic(ale), academic(kr)} 

\textit{student(oli), lecturer(ale), lecturer(kr)}
\end{itemize}

\noindent where the only observable predicate is \textit{sad}/1.\\

Given the observations \textit{O} = \{\textit{sad}(\textit{ale}), \textit{sad}(\textit{kr}), \textit{not}~\textit{sad}(\textit{oli})\} can we improve our model? The incompleteness of our model resides in the predicate \textit{poor}. This is the only abducible predicate in our model. Using abduction we can explain the observations \textit{O} via the explanation \textit{E = \{poor(ale), poor(kr), not poor(oli)}\}.

\noindent Subsequently, treating this explanation as training data for inductive generalization we can generalize this to get the rule:

\begin{itemize}
\item[] \textit{poor(X) \quad  $\leftarrow$ \quad  lecturer(X)}
\end{itemize}

\noindent thus (partially) defining the abducible predicate \textit{poor} when we extend our theory with this rule. In other words, we have used abduction to transform the training data from one level, the directly observable level, to training data at another level, the underlying abducible level, where the learning is carried out.
\end{example}

\subsubsection{Abduction in ILP Learning}

This combination of abduction and induction has been deployed and studied in several ways within the context of ILP. In particular, \textit{inverse entailment} \cite{MuggBryant} can be seen as a particular case of integration of abductive inference for constructing a ``bottom'' clause and inductive inference to generalize it. This is realized in Progol 5.0 and applied to several application problems. Similarly, an ILP system called ALECTO \cite{Moyle2000} integrates a phase of \textit{extraction-case abduction} to transform each case of a training example to an abductive hypothesis with a phase of induction that generalizes these abductive hypotheses. Other ILP frameworks based on the integration of abductive and inductive reasoning can be found in \cite{TAL_Russo,Russo2016,LAMMA1999}.

The development of these frameworks that realize the cycle of integration of abduction and induction prompted the study of the problem of \textit{completeness} for finding any hypotheses \textit{H} that satisfies the basic task of finding a consistent hypothesis \textit{H} such that $T \cup H\models O$ for a given theory \textit{T}, and observations \textit{O}. Progol was found to be incomplete \cite{yamamoto} and several new frameworks of integration of abduction and induction have been proposed such as SOLDR \cite{yamamoto:implementation}, CF-induction \cite{Inoue04}, and HAIL \cite{Ray1}. In particular, HAIL has demonstrated that one of the main reasons for the incompleteness of Progol is that in its cycle of integration of abduction and induction, it uses a very restricted form of abduction. Lifting some of these restrictions, through the employment of methods from abductive logic programming \cite{ALP}, has allowed HAIL to solve a wider class of problems. HAIL has been extended to a framework, called XHAIL \cite{RayXHAIL2009}, for learning nonmonotonic ILP, allowing it to be applied to learn Event Calculus theories for action description \cite{RayRusso2009} and complex scientific theories for systems biology \cite{RayBryant2008}.

\subsubsection{Abduction in Neural-Based Learning}

In the context of building Explainable AI systems, Abduction can help generate cognitive explanations for Machine Learning predictions. Unlike explanations over logic-based or symbolic learning methods (using, e.g., decision trees), which can be directly obtained from the model \cite{ignatiev2018sat,lakkaraju2016interpretable}, explanations over neural or sub-symbolic learning models (using, e.g., deep learning or Bayesian classifiers) are not naturally provided by the model, and need to be obtained by building a parallel symbolic model \cite{ignatiev2019abduction}.

In an orthogonal direction, a symbolic module can be built on top of a neural module so that data are fed into the neural module, whose outputs are fed, in turn, into the symbolic module. The latter, then, computes the final outputs, or predictions, of the integrated system, and these predictions are expected to match the high-level labels of the data. In this setting, explanations of the symbolic module take on the role of providing cognitively understandable lower-level labels for the training of the neural module; see, e.g., \cite{DeepAbduction,NeuroLog}.

It has been shown, in fact, that abduction supports a clean and compositional integration of the neural and symbolic modules, without imposing restrictions on their syntax and semantics \cite{NeuroLog}. Unlike previous neural-symbolic integration approaches that generally assume that the symbolic module encodes a theory that is effectively differentiable --- and, thus, compatible with the typical neural learning process of backpropagation --- abduction accommodates any theory for the symbolic module, and utilizes the theory's abductive explanations to compute a loss function that is itself differentiable, even if the theory is not.

It is then easy to see how the particular framework can support an abduction-induction cycle applied on the symbolic module as follows: During the abduction part of the cycle, we assume the symbolic module is fixed, and we abduce labels that we use to train the neural module. During the induction part of the cycle, we assume the neural part is fixed, and we deduce through it data to train the symbolic module. Thus, the cycle iteratively improves how the neural module maps low-level data to high-level data, on which to train the symbolic module.


\subsection{Argumentation in Machine Learning}

We have argued in Section~\ref{Section: Motivation and Background} that argumentation can offer a logical notion of coverage and prediction for learning that would naturally deal with the uncertainty, conflicts and incompleteness that are intrinsic characteristics of learning. Although there are other formulations which share this characteristic, often called non-monotonic logics, the foundational link of argumentation to logic-based learning comes from its advantage that learning can be directly related to building and generalizing arguments from the training data. Indeed, learning can be seen as uncovering the \emph{reasons} for why we observe certain phenomena, i.e., why the given training data is the way it is. We learn therefore arguments that can best support the observed data --- that help us classify the data one way instead of another in terms of arguments that support well the given data.

These reasons, or arguments, manifest themselves in various phases of the learning process, and this manifestation provides a dimension across which relevant work can be categorized. Orthogonally to this categorization, a key aspect that permeates all relevant work is the learning of priorities, or the defense relation, between arguments. Three are the main approaches followed: \cond{i} priorities derive from a notion of subsumption on the premises of arguments \cite{CBR, A-ART, MAICL1, MAICL2}; \cond{ii} priorities are determined by the learning data \cite{DK, EL, SLAP, NERD, SynthesizingAF, Nicoletta, DL}; or \cond{iii} priorities are unavailable or capture domain expert knowledge and are provided externally \cite{CLA, CleAr1, CleAr2, AARL1, AARL2, AARL3, MC1, MC2, ABML}. Additional considerations regarding a subset of the works reviewed below can be found in \cite{Toni_AMLsurvey}.

\subsubsection{Inputs Extended with Arguments}

A first line of work integrates argumentation with learning instances, as \emph{a way to enhance the information that is communicated to the learning process}.

In Argumentation-Based Machine Learning (ABML) \cite{ABML}, a learning instance comprises not only the input and output that specify, respectively, the features and the label of the instance, but also an associated logic-based argument that explains why the particular label is the case as a function of a subset of the features. Thus, upon seeing a small bean with a black color, the learning process also receives the argument ``black bean because small bean''.

Such side information offers a glimpse of the structure of the target concept that is being learned, which goes beyond what a single labeled learning instance would offer. Unsurprisingly, then, the process of learning benefits greatly in terms of performance, as demonstrated across domains \cite{ABML_zoology,ABML_law,ABML_medical}.

A similar in flavor approach is also taken in Argumentation-Accelerated Reinforcement Learning (AARL) \cite{AARL1,AARL2,AARL3}, where the arguments offered provide conditions under which a particular action should be taken in a Reinforcement Learning context. Such arguments can help shape the reward, so that the learning process can then more efficiently learn a good policy.

Extending the basic idea of accompanying learning instances with applicable arguments, Machine Coaching \cite{MC1,MC2} takes the view that these arguments are contextualized on the current state of the learning process, and are not, therefore, available up front. Rather, learning proceeds in an online fashion, where the input of the learning instance is first presented, the learning process makes a prediction on the instance's label, explains how that prediction came about, and only then receives an argument in response to that explanation. The argument itself need not argue with respect to the prediction itself, but could also argue about intermediate concepts that appear in the offered explanation.

When observing a small bean drawn from the bag, for example, the prediction could be that ``normal bean because drawn from the bag'' and ``white bean because normal bean'', while the counterargument could be that ``not normal bean because small bean''. Formal analysis of this protocol shows that efficient learning is indeed facilitated by such counter-arguments, even in cases where efficient learning from input-output instances alone would not be possible.

\subsubsection{Inputs Interpreted as Arguments}

A second line of work integrates argumentation with the learning process, as \emph{a way to interpret learning instances in case of noisy or partial information}.

Concept Learning as Argumentation (CLA) \cite{CLA} takes the view that each (noisy) learning instance is an argument that states that for that particular input, the corresponding output is the appropriate instance label. Such arguments are taken to be strong, or preferred, as they derive directly from \emph{observed} information. At the same time, each potential hypothesis among those that could result through learning, offers its own argument on the label of each learning instance, stating that for that particular input, the hypothesis prediction is the appropriate instance label. Such arguments are taken to be weak, as they are premised on the \emph{assumption} that the hypothesis is accurate. Resolving the conflicts between the arguments to compute an acceptable extension effectively amounts to a process of learning, with the extension being the learned outcome.

Restricting attention to only the strong arguments from above, one effectively ends-up with the case-based reasoning (CBR) framework in \cite{CBR}, where each learning instance offers a case in support for its corresponding label. To predict the label of a new learning instance, then, one seeks to find which among the previous cases is closest to the new learning instance in terms of their shared features; such a closest case would then provide an argument for the label of the new learning instance to be the same as the label of that case.

By observing a white bean from the bag and a small black bean from the bag, for example, one ends up with the two arguments ``white bean if from the bag'' and ``black bean if small bean from the bag''. Cases (or argument conditions) are not disjoint, and thus multiple cases might be relevant when trying to predict the label of a new learning instance. Argumentation over those cases resolves which cases are to be considered, by offering priority to cases that are more specific: the small bean from the bag, in this example.

Instead of considering individual learning instances as arguments, one can consider arguments derived from subsets of instances. These arguments can, as before, be reasoned with to determine which ones are acceptable, and hence, which ones will be used to support a prediction on a new learning instance. Argumentation for Multi-Agent Inductive Concept Learning (MAICL) \cite{MAICL1,MAICL2} approaches this setting by thinking of the subsets of instances as data split among different agents, and by thinking of the corresponding arguments as the hypotheses that were learned by the agents from their given subset of data. In this distributed learning setting, then, argumentation aims to reconcile the different hypotheses that were learned by the agents, so that they can collectively decide the prediction on a new learning instance.

\subsubsection{Hypotheses Interpreted as Arguments}

Beyond integrating argumentation during the learning process to enhance or interpret learning instances, a third line of work utilizes argumentation after the learning process, as \emph{a way to contrast the learned hypothesis against competing learned hypotheses or external alternatives}.

Classification enhanced with Argumentation (CleAr) \cite{CleAr1,CleAr2} considers a setting where expert opinion or domain knowledge on how a new input should be classified might be available, and these alternatives need to compete against or support the learned hypothesis and each other. Argumentation helps resolve the conflicts that arise, with each argument being assigned a base score, and these scores being used to decide which conclusion prevails.

A similar situation can be observed in the context of unsupervised learning, where one ends up with clusters of training inputs, with each of them effectively providing a competing alternative on how a new input should be classified (i.e., in which cluster it should belong). Argumentation for ART (A-ART) \cite{A-ART} uses argumentation to contrast the competing learned clusters and decide which among them to employ, with priorities given to the various clusters based on a subsumption relation that might exist between them.

MAICL \cite{MAICL1,MAICL2}, as introduced earlier, can also be seen to fall into this group of approaches, since each agent ends up with a competing learned hypothesis.

\subsubsection{Hypotheses Expressed in Argumentation}

In the preceding approaches, argumentation was used to provide semantics on how a learned hypothesis interacted with other entities, those being either the learning inputs, or other learned hypotheses or alternative classifiers. The internal structure and semantics of the learned hypothesis was not, however, necessarily specified. In some of the previously described approaches, in fact, hypotheses were, or could be, learned using standard learning algorithms with their associated hypothesis spaces. On the other hand, a body of work has considered argumentation to be the target language for learning, as \emph{a way to specify the representation and reasoning semantics of a learned hypothesis}. 

One line of work \cite{SynthesizingAF} has considered the problem, and the complexity, of learning an abstract argumentation framework, and hence effectively the attack relation between the arguments, by considering as learning inputs extensions and non-extensions under various argumentation semantics. A number of other works have considered the problem of learning a structured argumentation framework, where the arguments have internal structure, which itself needs to be learned in addition to the attack relation that holds between them.

Work on learning Decision Lists \cite{DL} and Exception Lists \cite{EL} can be seen to fall in the latter group of works, since such prioritized lists effectively totally order a set of competing conditions as a way to resolve conflicts between their conclusions. In both cases, learning proceeds under Valiant's PAC semantics \cite{PAC}. Another work \cite{DK} approaches the problem under Gold's learning in the limit semantics \cite{ITL}, and shows how learning rules and exceptions can be done iteratively, as a way to systematically enhance the coverage of the learned hypothesis. Priorities between rules effectively correspond to the specificity of rule conditions. Finally, some recent work \cite{Nicoletta} constructs first an interpretable hypothesis through a standard learning algorithm (random forests, in particular), and subsequently extracts rules from that learned model, which are used as arguments along with additionally learned conditional priorities between them.

The works above assume that arguments comprise individual rules that directly map their inputs to the supported conclusions. Another body of work has sought to learn argumentation theories with arguments comprising multiple chained rules that map inputs to conclusions through intermediate concepts.

The Never-Ending Rule Discovery (NERD) algorithm \cite{NERD} offers a heuristic to learn rules in an online fashion where inputs are received in a streaming fashion. In that work, priorities between rules are determined not by the structure of the rules themselves, but are dictated by the learning data, since rules that are found to be sufficiently supported by the data first are given higher priorities. Machine Coaching \cite{MC1,MC2} has taken an alternative approach, where a human coach reacts to the current learned hypothesis and the arguments that are offered in support of a given input, and provides counter-arguments as feedback to enhance the learned hypothesis. Priorities are given based on the recency of the arguments, where more recently-provided counter-arguments defeat earlier ones. Finally, work on Simultaneous Learning and Prediction (SLAP) \cite{ALR,SLAP,EKI} learns rules in a stratified manner, with priorities between rules, and hence between the resulting arguments, derived from this stratification. Both Machine Coaching \cite{MC1,MC2} and SLAP \cite{ALR,SLAP,EKI} adopt the PAC semantics \cite{PAC}.


\section*{Acknowledgements}

This work was supported by funding from the EU's Horizon 2020 Research and Innovation Programme under grant agreements no.\ 739578 and no.\ 823783, and from the Government of the Republic of Cyprus through the Directorate General for European Programmes, Coordination, and Development.


\bibliographystyle{plain}
\bibliography{references}

\end{document}